\documentclass[letterpaper, 10pt, conference]{ieeeconf}
\usepackage{hyperref} 

\IEEEoverridecommandlockouts                              

\usepackage{graphicx}
\usepackage[usenames,dvipsnames]{xcolor}
\usepackage[fleqn]{amsmath}
\usepackage{amssymb}
\usepackage{amsfonts}
\usepackage{mathrsfs}
\usepackage{acronym}
\usepackage{float}
\usepackage{epsfig}
\usepackage{cite}
\graphicspath{{Figures/}}
\usepackage{url}
\usepackage{stfloats}
\usepackage{amsopn}     
\usepackage{psfrag}
\usepackage{subfigure}
\usepackage{mdwtab}     
\usepackage{commath}
\usepackage{enumerate}
\usepackage{balance}
\usepackage{bm}
\usepackage{commath}
\usepackage{paralist}
\usepackage{acronym}
\usepackage{todonotes}

\DeclareMathOperator{\erf}{erf}
\acrodef{lrt}[\textsc{lrt}]{Likelihood Ratio Test}
\acrodef{snr}[\textsc{snr}]{signal-to-noise-ratio}
\acrodef{imu}[\textsc{imu}]{inertial measurement unit}
\acrodef{gps}[\textsc{gps}]{global positioning system}
\acrodef{sde}[\textsc{sde}]{stochastic differential equation}
\acrodef{ou}[\textsc{ou}]{Ornstein-Uhlenbeck}
\acrodef{dspp}[\textsc{dspp}]{doubly stochastic Poisson process}
\acrodef{smp}[\textsc{smp}]{stochastic maximum principle}
\acrodef{fbsdes}[\textsc{fbsde}s]{forward-backward stochastic differential equations}
\acrodef{pde}[\textsc{pde}]{partial differential equation}
\acrodef{cps}[\textsc{cps}]{counts per second}
\acrodef{cots}[\textsc{cots}]{commercial-of-the-shelf}
\acrodef{uav}[\textsc{uav}]{unmanned aerial vehicle}
\acrodef{msckf}[\textsc{msckf}]{multi-state constraint Kalman filter}
\acrodef{vins}[\textsc{vins}]{visual-inertial navigation systems}
\acrodef{ekf}[\textsc{ekf}]{Extended Kalman Filter}
\acrodef{rmse}[\textsc{rmse}]{root mean square error}

\DeclareGraphicsExtensions{.eps,.pdf,.png,.jpg,.JPG,.gif}

\renewcommand{\baselinestretch}{.98}
\title{\LARGE \bf
Visual-Inertial Target Tracking and Motion Planning for UAV-based Radiation Detection
}

\author{Indrajeet Yadav*, Kevin Eckenhoff*, Guoquan Huang, and Herbert G. Tanner
\thanks{*These authors contributed equally to this work.}
\thanks{This work was partially supported by DTRA (HDTRA1-16-1-0039).}%
\thanks{The authors are with the Department of Mechanical Engineering,   University of Delaware, Newark, DE 19716. Email:
        {\tt\small \{indragt, keck, ghuang, btanner\}@udel.edu}}%
}

\begin{document}

\pagenumbering{gobble}

\maketitle

\thispagestyle{empty}
\pagestyle{empty}



\begin{abstract}

This paper addresses the problem of detecting radioactive material in transit using an \ac{uav} of minimal sensing capability,
where the objective  is to classify the target's radioactivity as the vehicle plans its paths through the workspace  while tracking the target for a short time interval. 
To this end, we propose a motion planning framework  that integrates tightly-coupled visual-inertial localization and target tracking. 
In this framework,
the 3D  workspace is \emph{known}, and this information together with the \ac{uav} dynamics, is used to construct a navigation function that generates dynamically feasible, safe paths which avoid obstacles and provably converge to the moving target.
%
The efficacy of the proposed approach is validated through realistic simulations in Gazebo. 

\end{abstract}

\section{Introduction}\label{Intro}
%
%

The ultimate goal of this work is to identify whether a  target moving in a cluttered environment is radioactive or not, in a matter of a few minutes, using an airborne radiation sensor (e.g.~a Geiger counter or Neutron detector).
As the potential radioactivity of the target is assumed to be known, the problem essentially is an instance of binary hypothesis testing -- 
{\em does the target have the assumed intensity or not?} 
Earlier work \cite{Sun} has established that the optimal way, in a Neyman-Pearson sense \cite{PSPT_Automatica}, of answering this question using a controllable mobile radiation detector is to drive the detector as fast as possible to the moving target.
Intuitively, this strategy maximizes over time the solid angle of the radiation sensor, boosting its \ac{snr} \cite{Nemzek}, thus improving the accuracy of decision-making.

The ability of a \ac{uav} equipped with a radiation detector to complete this task autonomously in a cluttered and possibly \textsc{gps}-denied environment, rests on two critical capabilities: 
\begin{inparaenum}[(i)]
\item  estimating consistently and accurately   its own state and that of its target, and 
\item planning motion in a way that maximizes the probability of accurate detection while avoiding collisions with obstacles and the target.
\end{inparaenum}
In \emph{isolation}, both of these technical problems have received some attention in literature.

In the context of target tracking, most methods rest on assumptions that limit their applicability to real-world scenarios. 
For instance, many algorithms exist for robot-based target tracking in 2D~\cite{Huang2014RAS, Mirzaei2007ICRA}, without straightforward extensions to 3D cases.
One idea explored in the context of active tracking~\cite{ZhouKe2011TRO} is to choose robot motion in ways that minimize future target uncertainty. 
This methodology, again, was developed for 2D problems and requires that the state of the sensor is perfectly known. 
Other approaches assume the ability to directly measure the relative position between sensor and target~\cite{Chen2016IROS}, and estimate robot and target states separately; because of this separation, however, there is little that can be said about the optimality of either estimate, or a quantitative measure of the \textit{uncertainty} associated with them.
Compounding these technical challenges in the application considered here, the (unknown) relative distance between sensor and target is a significant source of uncertainty when it comes to the accuracy of the final, radiation classification decision that needs to be made.

The proposed approach  \textit{tightly} couples 3D visual-inertial navigation and target tracking, to ensure that the error in estimating the relative distance between the robot and the target can be minimized. 
In the application at hand, this is particularly important because this relative distance affects directly the information content of radiation measurements, and consequently the radiation detection accuracy. 
Estimating the relative position between detector and target is achieved with exclusive use of an \ac{imu} and a (stereo) camera, on the hardware side, and a lightweight \ac{msckf}~\cite{Mourikis2007ICRA} on the algorithmic side. 
The particular estimator utilizes stochastic cloning to estimate a sliding window of past robot poses along with the current \ac{imu} state. 
Features are linearly marginalized to utilize their motion information without the need to store them in the state vector. The marginalization allows for the creation of constraints between the window poses while bounding the problem size, resulting in a computationally efficient estimation algorithm. 
This approach has been extended in many directions, e.g., including camera-to-\ac{imu} spatial and temporal calibration~\cite{Li2014IJRRa}, handling degenerate motions~\cite{Kottas2013IROS}, and enforcing the correct observability properties on the system~\cite{Hesch2013TRO}. 

Aerial target tracking and navigation based on-board estimation has been recently demonstrated in a case of tracking a spherical rolling target~\cite{VK1}.
It employed a geometric approach similar to visual servoing, a receding horizon strategy that penalizes velocity and position errors, and a \ac{uav} motion control scheme based on minimum-snap trajectory generation~\cite{Mellinger2011ICRA}.
It is not clear to what degree this tracking  algorithm depends on the target being spherical, but the overall estimation and motion control scheme was engineered to run impressively fast.


The work reported here involves an additional layer of complexity.
In addition to target tracking and obstacle avoidance, the mission calls for fast and accurate target classification based on a different sensing modality.
Here target classification, estimation, and tracking are coupled; because the accuracy of estimates and the performance of tracking have direct impact on the accuracy of classification.
The size of tracking errors reported in~\cite{Chen2016IROS} can be detrimental to detection of low-intensity mobile sources of radioactivity.

In the application context of radiation detection, (binary) classification accuracy can be quantified probabilistically in terms of probabilities of \emph{false alarm} and \emph{missed detection}.
The computation of the latter two probabilities are in general intractable; however, analytical Chernoff bounds on these have recently been derived~\cite{PSPT_Automatica} in the form of highly non-convex, non-linear integral functions.
While these analytic expressions provide a handle for manipulating accuracy of decision-making through mobility, they are still not particularly amenable to direct incorporation into standard motion planning and optimal estimation frameworks. 
Yet we do know \cite{Sun} that in the absence of constraints in terms of vehicle dynamics or workspace topology, the optimal motion strategy for maximizing detection accuracy is to close the distance between radiation detector and source as quickly as possible.

This \emph{insight}, therefore, is what can be utilized by means of a variety of motion planning strategies \cite{RRT}.
Not all of them, however, can offer \emph{provable} collision avoidance and target convergence guarantees, especially when considering nontrivial sensor platform dynamics.
The approach in this paper achieves this by combining a new type of navigation functions developed for moving targets \cite{Sun}, with a geometric position and attitude tracking \ac{uav} controller.
The navigation function creates an almost globally attractive (to the target) vector field, which is utilized as a velocity reference for the \ac{uav}'s controller.
Vector field tracking has been entertained in earlier work~\cite{Schwager}, but neither for the purpose of intercepting moving targets, nor in non-spherical workspace topologies and without full knowledge of robot and target states~\cite{Roussos}.

In particular, the main  contribution of this paper consist of
\begin{inparaenum}[(i)]
\item the extension of \ac{vins} to tightly-coupled 3D visual-inertial localization and target tracking 
\item integration of geometric \ac{uav} control with time-varying navigation functions in star-forests.
\item validation of the proposed system in realistic {\small \textsf{ROS/Gazebo}} simulations.
\end{inparaenum}

\section{Radiation Detection}\label{detection}

To a radiation detector, gamma rays/neutrons emanating from nuclear material are indistinguishable from naturally occurring (background) radiation. 
In the sensor's input stream, the two signals are essentially superimposed. 
To (optimally) determine whether a source is present based on measurements from such a radiation counter, one performs a binary hypothesis test. 

Pahlajani et al.~\cite{PSPT_Automatica}  formulate this problem as a \ac{lrt} in the Neyman-Pearson framework. 
In this framework, one of the following competing hypotheses is chosen: $H_0$, which asserts that the target is benign, or $H_1$, which states that the target is radioactive.
In making this decision, two types of errors can occur. 
False alarm occurs if the target is falsely classified as radioactive; missed detection occurs when the target is falsely classified as benign. 

Analytically computing the probability of making either error is in general intractable, except for (simplistic) Gaussian counting processes. 
Not only is the counting process here Poisson, but it is also a time-inhomogeneous one, since it explicitly depends on the distance between sensor and source which changes (deterministically) with time. 
Yet, analytical Chernoff (upper) bounds for the errors involved in this particular binary hypothesis test have been derived~\cite{PSPT_Automatica}, and these can serve as surrogates for the otherwise unknown true error probabilities.

Let $N_t$ for $t \in [0,T]$ be the time-inhomogeneous counting process observed at a detector with cross-section $\chi$ (square meters), when monitoring a source of activity $a$ in \ac{cps} in an environment with background radiation of activity $b(t)$.
The quantities ${}^\mathsf{M}\mathbf{p}(t)$ and $ {}^\mathsf{M}\mathbf{p}_T(t) \in \mathbb{R}^3$ denote the position of the detector (hereafter assumed identical to that of the \ac{uav}) and the target, respectively, in the inertial \textsf{M}\textsc{ap} 
frame. 
The \emph{perceived} (given the source-sensor distance) source activity, incident to the detector is denoted $\nu(t)$, while a dimensionless parameter $\mu$ quantifies the relative strength of this perceived activity over background and is conceptually related to \ac{snr} (norms are Euclidean)
%
%
\begin{equation}\label{Eq:mu}
\nu(t) = \frac{\chi \, a}{2 \chi + \|{}^\mathsf{M}\mathbf{p}(t)-{}^\mathsf{M}\mathbf{p}_T(t)\|^2} \enspace, \quad    
\mu(t) = 1 + \frac{\nu(t)}{\mathit{b}(t)}
\end{equation}
Let $\alpha$ be an upper bound on the acceptable probability of false alarm, and $p\in(0,1)$ a constant parameter.
Then after defining
\begin{subequations}
\label{Eq:position}
\begin{equation}
\label{Eq:threshold}
\Lambda^{'}(p)=  \int_{0}^{T} [\mu^p \log \mu - \mu+1] \mathit{b} \dif s
\end{equation}
the logarithm of the analytically derived bounds on the probabilities of false alarm and missed detection is~\cite{PSPT_Automatica}
\begin{align}
\mathcal{E_F} &=  \int_{0}^{T} 
	[p\,\mu^{p} \log \mu - \mu^{p}+1] 
    \mathit{b} \dif s = -\log \alpha \label{Eq:PF}\\
\mathcal{E_M} &= \log \alpha + \Lambda^{'}(p) \label{Eq:PM} 
\end{align}
\end{subequations}
Parameter $p$ is specified by solving \eqref{Eq:PF} for a given $\alpha$. 

For $n \ge 1$, let $\tau_n$ denote the $n$-th jump time (when the detector registers a count) in a counting random process taking the form of piecewise constant random signal with maximum $N_t$ over a finite time window of $[0,T]$. 
The \ac{lrt} to decide between the two hypotheses $H_0$ and $H_1$ involves comparing a statistic $L_T$ to a constant threshold $\gamma = \exp(\Lambda^{'}(p)) \in \mathbb{R}$
\begin{align}\label{Eq:LT}
\hspace{-5ex}
\exp\left(-\int_{0}^{T} \nu (s) \dif s\right) \prod_{n=1}^{N_t}  \Bigg(1+ \frac{\nu (\tau_n)}{\mathit{b} (\tau_n)}\Bigg) =: L_T \mathop{\gtrless}_{H_0}^{H_1} \gamma  
\end{align}
The bound on probability of missed detection \eqref{Eq:PM} serves as a performance measure of decision-making accuracy when performing the \ac{lrt}.

\section{Motion Planning}

\subsection{Navigation Function}\label{planning}

A navigation function~\cite{Rimon} is a real-valued map $\mathcal{V}: \mathcal{F} \rightarrow \mathbb{R}$, constructed on \ac{uav}'s free configuration space  $\mathcal{F}$ that when tuned appropriately has a unique minimum at the desired goal configuration and is uniformly maximal over the boundary of $\mathcal{F}$. 
The construction has been extended to the case of goal manifolds that take the form of a sphere (bubble) around a moving target~\cite{Sun}.
Because these moving manifolds are (almost) globally attractive, the use of this motion planning technique guarantees the convergence of the \ac{uav} to the moving target within a specified distance, without hitting either this target or the surrounding obstacles.




Let ${}^\mathsf{M}\mathbf{p}(t)$ denote the position of the \ac{uav}, and ${}^\mathsf{M}\mathbf{p}_T(t)$ that of the target, assumed both in $\mathcal{F}$. 
Take $r$ to be the radius of the spherical bubble around the target and define the goal for navigation as minimizing the function
%
\begin{equation}
J\big( 
	{}^\mathsf{M}\mathbf{p}, {}^\mathsf{M}\mathbf{p}_T 
\big) = 
\|{}^\mathsf{M}\mathbf{p} - {}^\mathsf{M}\mathbf{p}_T\|^2 - r^2
\label{Eq:goal_func}
\end{equation}
Let $\{0,\ldots,S\}$ be the index set of spherical obstacles, including the workspace's outer boundary indexed $0$, and denote $o_i$, $\rho_i$ for $i \in \{0,\ldots,S\}$ the center and radius of each spherical obstacle, respectively.
Define 
$\beta_0({}^\mathsf{M}\mathbf{p}, {}^\mathsf{M}\mathbf{p}_T) = \rho_0^2-\|{}^\mathsf{M}\mathbf{p}-o_0\|^2$
for the workspace boundary, and
$\beta_i({}^\mathsf{M}\mathbf{p}, {}^\mathsf{M}\mathbf{p}_T) = \|{}^\mathsf{M}\mathbf{p}-o_i\|^2 - \rho_i^2$ for any other interior obstacle.
Let
$\beta({}^\mathsf{M}\mathbf{p}, {}^\mathsf{M}\mathbf{p}_T) = \prod_{i = 0}^{S} \beta_i({}^\mathsf{M}\mathbf{p}, {}^\mathsf{M}\mathbf{p}_T)$. 

It can be shown~\cite{Sun} that there exists a positive number $N$ such that $\forall k \geq N$, and for a suitably large parameter $\lambda \in \mathbb{R}_+$,   
%
%
\begin{equation}
\tilde{\varphi} 
({}^\mathsf{M}\mathbf{p}, {}^\mathsf{M}\mathbf{p}_T) 
= 
\frac
{J ({}^\mathsf{M}\mathbf{p}, {}^\mathsf{M}\mathbf{p}_T)}
{\big[
	J ({}^\mathsf{M}\mathbf{p}, 
    	{}^\mathsf{M}\mathbf{p}_T)^\kappa 
   + \lambda \, 
   \beta(
   	{}^\mathsf{M}\mathbf{p}, 	
   	{}^\mathsf{M}\mathbf{p}_T
   )
 \big]^{1/\kappa}}
\label{Eq:sp_nav_func}
\end{equation}
would be a navigation function on a spherical world $\mathcal{S}$ with a suitably chosen (analytic switch) parameter $\lambda$. 
%
%

For a diffeomorphism $h_{\lambda_{\mathrm{sq}}}$ parameterized by a suitably chosen positive parameter $\lambda_{\mathrm{sq}} \in \mathbb{R}_+$, 
mapping a star world $\mathcal{F}$ to the forest of stars in the form of three-dimensional \emph{squircles}~\cite{caili},
the composition 
$ 
\varphi = \tilde{\varphi} \circ h_{\lambda_{sq}}
$
can be shown to have navigation function properties on $\mathcal{F}$~\cite{caili},  in the sense that for any position of the target satisfying some reasonable conditions, all (unstable) critical points outside the destination manifold are either nondegenerate with attraction region of measure zero, or inside the target's bubble.



\subsection{\textsc{Uav} Control}\label{dynamics}

Let $m$ denote the mass of the quadrotor \ac{uav}, $\mathbf{J} \in \mathbb{R}^{3 \times 3}$ its moment of inertia about a frame aligned with the principal axes and attached at the center of mass $\mathsf{cg}$, and $^\mathsf{M}\bm{g}$ the acceleration of gravity in the inertial  frame.
The relative orientation between the inertial frame and the principal one at the \ac{uav}'s center of mass is captured by the rotation matrix ${}^\mathsf{M}_\mathsf{cg} \mathbf{R} \in \mathsf{SO}(3)$.
(Premultiplication with ${}^\mathsf{M}_\mathsf{cg} \mathbf{R}$ rotates a vector by as much as \textsf{cg} is rotated relative to \textsf{M}.)
The linear and angular velocity of the \ac{uav} relative to the inertial frame are denoted ${}^\mathsf{M}\mathbf{v}$ and ${}^\mathsf{M} \bm \omega$, respectively; the angular velocity vector relative to the body \textsf{cg} frame is denoted ${}^\mathsf{cg}\bm \omega$.
Let $\widehat{\cdot}$ denote the (wedge) operation that maps a vector in $\mathbb{R}^3$ to a member of the Lie algebra $\mathfrak{so}(3)$.
With $f$ denoting the magnitude of the total thrust produced by the thrusters, and $\mathbf{M}$ the total moment relative to the body-fixed \textsf{cg} frame, the dynamics of the quadrotor can finally be expressed as
%
\begin{subequations}
\label{uav-dynamics}
\begin{align}
{}^\mathsf{M}\dot{\mathbf{p}} &= {}^\mathsf{M}\mathbf{v}\\
m\; {}^\mathsf{M}\dot{\mathbf{v}} &= -m\; {}^\mathsf{M}\mathbf{g} + f\; {}^\mathsf{M}_{\mathsf{cg}}\mathbf{R}\;\tfrac{ ^\mathsf{M}\mathbf{g}}{\| ^\mathsf{M}\mathbf{g} \|}\\
{}^\mathsf{M}_{\mathsf{cg}}\dot{\mathbf{R}} &= {}^\mathsf{M}_{\mathsf{cg}}\mathbf{R} \;  \widehat{ {}^{\mathsf{cg}} \bm \omega }\\
\mathbf{J}\; {}^{\mathsf{cg}}\bm \omega +  {}^{\mathsf{cg}}\bm \omega \times  \mathbf{J} \;{}^{\mathsf{cg}}\bm \omega &= \mathbf{M}
\end{align}
\end{subequations}

The desired velocity for the \ac{uav} is determined using \eqref{Eq:sp_nav_func}.
Specifically, if $v_\mathrm{max}$ denotes the vehicle's maximum speed given the capabilities of its actuators or safety specifications, and $k$ is a positive control gain, then the velocity reference relative to the inertial frame is 
\begin{equation}
{}^\mathsf{M}\mathbf{v}_d \triangleq - \erf{\big( k  (\|{}^\mathsf{M}\mathbf{p}-{}^\mathsf{M}\mathbf{p}_T\|-r)\big)} \cdot \frac{\nabla_\mathbf{p} \varphi}{\|\nabla_{\mathbf{p}} \varphi\|} \cdot v_\mathsf{max}
\label{Eq:control_action}
\end{equation}
Consequently, the velocity error for the \ac{uav} would simply be $\mathbf{e}_v \triangleq {}^\mathsf{M}\mathbf{v} - {}^\mathsf{M}\mathbf{v}_d$.
With ${}^\mathsf{M}_\mathsf{cg}\mathbf{R}_d$ and $\bm \omega_d$ denoting some desired \ac{uav} orientation and angular velocity for the \ac{uav}, respectively,
the orientation and angular velocity errors are defined accordingly 
\begin{subequations}
\label{Eq:error2}
\begin{align}
\widehat{ \mathbf{e}_R }  &= \tfrac{1}{2} \big( {}^\mathsf{M}_{\mathsf{cg}}\mathbf{R}^\intercal_d \;{}^\mathsf{M}_\mathsf{cg}\mathbf{R}-{}^\mathsf{M}_\mathsf{cg}\mathbf{R}^\intercal \; {}^\mathsf{M}_\mathsf{cg}\mathbf{R}_d \big)\\
\mathbf{e}_{\omega} &= {}^\mathsf{cg} \bm \omega - {}^\mathsf{M}_\mathsf{cg}\mathbf{R}^\intercal \; {}^\mathsf{M}_\mathsf{cg}\mathbf{R}_d \; {}^\mathsf{cg}\bm \omega_d
\end{align}
\end{subequations}
With these definitions in place, and with $k_v$, $k_R$ and $k_\omega$ denoting positive control gains for velocity, orientation, and angular velocity, the control inputs for the \ac{uav} are set as (cf.~\cite{Lee})
%
%
\begin{subequations}
\label{uav-controls}
\begin{align}
f &= \big( -k_v \mathbf{e}_v + m\; {}^\mathsf{M}\mathbf{g}  + m \; {}^\mathsf{M}\dot{\mathbf{v}_d} \big) \cdot 
\big( \, {}^\mathsf{M}_\mathsf{cg}\mathbf{R} \; \tfrac{^\mathsf{M} \mathbf{g} }{\| ^\mathsf{M}\mathbf{g} \|} \, \big) 
\label{Eq:dynamics1}
\\
\mathbf{M} &= -k_R \, \mathbf{e}_R - k_{\omega} \, \mathbf{e}_{\omega} +   {}^\mathsf{cg} \bm \omega \times  \mathbf{J} \; {}^\mathsf{cg} \bm \omega \label{Eq:dynamics2}
\end{align}
\end{subequations}
%
%
In the closed-loop system \eqref{uav-dynamics}--\eqref{uav-controls}, the errors $\mathbf{e}_v$, $\mathbf{e}_R$, and $\mathbf{e}_\omega$ converge exponentially to zero~\cite{Lee}.

\section{State Estimation}

\subsection{State Definition}

Both control, as well as detection, rely on accurate estimates of the states of the \ac{uav} and its target. 
%
It is assumed that the \ac{uav} is only equipped with an \ac{imu}, which provides linear acceleration and angular velocity readings, and a set of stereo cameras. 
The camera images provide bearing measurements to three classes of objects: the target, known landmarks (which can be obtained, e.g., through a previous mapping session), and unknown features. 
To deal with the limited computational resources on the \ac{uav}, and because low-latency is required, we extend a popular filtering-based \ac{vins} solution, the \ac{msckf}~\cite{Mourikis2007ICRA} to estimate the \ac{uav} and target states, and the transformation between the \ac{uav} and map frames, through fusion of inertial (\ac{imu}) and visual (camera) data. 

Typically within a visual-inertial localization framework, the state of the \ac{uav}, parameterized by that of the \ac{imu}, is represented by an element $\mathbf{x}_\textsc{imu} \in \mathsf{SU}(2) \times \mathbb{R}^{12}$, where $\mathsf{SU}(2)$ is the group of unit quaternions.
This state representation for the \ac{uav}'s \ac{imu} at time step $k$, with $\mathsf{G}$ denoting the \ac{imu}'s global frame and $\mathsf{I}_k$ the \ac{imu}'s local frame at step $k$, can be written explicitly as
 \begin{align}
     \mathbf{x}_{\textsc{imu},k} 
     = \begin{bmatrix} {{}_\mathsf{G}^{\mathsf{I}_k}\bar{\mathbf{q}}}^\intercal 
     & {\mathbf{b}_{w_k}}^\intercal 
     & {{}^\mathsf{G}\mathbf{v}_k}^\intercal 
     & {\mathbf{b}_{a_k}}^\intercal 
     & {{}^\mathsf{G}\mathbf{p}_k}^\intercal 
     \end{bmatrix}^\intercal
 \end{align}
which includes the unit quaternion parameterizing the rotation from the global frame to the local \ac{imu} frame, gyro bias, global velocity, accelerometer bias, and global position, respectively. 
The convention for quaternions and their corresponding rotation matrices is that the left subscript denotes the starting frame of the rotation, while the left superscript denotes the end frame. 
When used with vector quantities, the left superscript indicates the frame in which the vector is being represented relative to.

The state of the target, $\mathbf{T} \in \mathbb{R}^{3n+3}$ is parameterized using a constant derivative model of order $n$, in which ${}^\mathsf{G}\mathbf{p}^{(i)}_{T_k}$ denotes the derivative of the target's position of order $i$,
\begin{align*}
 &\mathbf{T}_k = \begin{bmatrix} {{}^\mathsf{G}\mathbf{p}_{T_k}}^\intercal 
 & {{}^\mathsf{G}\mathbf{p}^{(1)}_{T_k}}^\intercal  
 & {{}^\mathsf{G}\mathbf{p}^{(2)}_{T_k}}^\intercal  
 & \cdots 
 & {{}^\mathsf{G}\mathbf{p}^{(n)}_{T_k}}^\intercal  
 \end{bmatrix}^\intercal
 \end{align*}
 
Static quantities also need to be estimated:  
$
\mathbf{x}_\mathsf{static} = \begin{bmatrix} 
\mathbf{x}_\mathsf{clones}^\intercal 
& 
\mathbf{x}_\textsc{map}^\intercal\end{bmatrix}^\intercal
$.
 The first term denotes a set of stochastic clones of the past $N$ imaging time \ac{imu} poses
\begin{align*}
 \mathbf{x}_\mathsf{clones} 
 &= \begin{bmatrix}\mathbf{x}_\mathsf{clone,1}^\intercal 
 & \mathbf{x}_\mathsf{clone,2}^\intercal 
 & \cdots 
 & \mathbf{x}_\mathsf{clone,N}^\intercal 
 \end{bmatrix}^\intercal \\
  \mathbf{x}_\mathsf{clone,i} 
  &= \begin{bmatrix} 
  {{}_\mathsf{G}^{\mathsf{I}_{i}}
  \bar{\mathbf{q}}}^\intercal 
  & {{}^\mathsf{G}\mathbf{p}_{i}}^\intercal
  \end{bmatrix}^\intercal 
  \in \mathsf{SU}(2) \times \mathbb{R}^3
\end{align*}
 These clones are maintained as per the \ac{msckf} framework, wherein bearing measurements to unknown features are transformed into relative constraints between historical poses (see Section~\ref{sec:update}). 
The second static term refers to the transformation from the \ac{imu}'s ``global'' frame, $\mathsf{G}$, and the frame of the prior map, $\mathsf{M}$. 
This term is required as the estimator initializes its own frame during startup \textit{without} any knowledge of its pose relative to the prior map. 
With ${}^\mathsf{G}\mathbf{p}_\mathsf{M}$ denoting the prior map's origin  in the global frame,
the relative transformation between frames is parameterized by
\[
  \mathbf{x}_\textsc{map} 
  = \begin{bmatrix} 
  {{}_\mathsf{G}^\mathsf{M}\bar{\mathbf{q}}}^\intercal 
  & {{}^\mathsf{G}\mathbf{p}_\mathsf{M}}^\intercal
  \end{bmatrix}^\intercal \in 
  \mathsf{SU}(2) \times \mathbb{R}^3
\]

The full state of this system vector at time step $k$ is now
\[
\mathbf{x}_k = \begin{bmatrix}
{\mathbf{x}_{\textsc{imu},k}}^\intercal 
& 
{\mathbf{T}_{k}}^\intercal 
& 
{\mathbf{x}_{\mathsf{static},k}}^\intercal
\end{bmatrix}^\intercal 
\]
It is important to note that the true state, $\mathbf{x}$, does not belong to a vector space, making direct estimation of it difficult with standard techniques. Instead, as in many recent \ac{vins} algorithms, these quantities have been \textit{indirectly} estimated  by performing filtering on a minimal representation error state vector. 
The relationship between the true value of the state, $\mathbf{x}$, the estimated state, $\hat{\mathbf{x}}$, and error state, $\delta{\mathbf{x}}$, is defined by the generalized update operation $\mathbf{x}= \hat{\mathbf{x}} \boxplus \delta{\mathbf{x}}$. For vector terms, this operation takes the form $\mathbf{x}= \hat{\mathbf{x}}+\delta{\mathbf{x}} $;  quaternion errors, with $\otimes$ representing quaternion multiplication~\cite{Trawny2005_Q_TR}, are represented as $\bar{\mathbf{q}}= \hat{\bar{\mathbf{q}}} \boxplus \delta{\bm \theta} = \delta \bar{\mathbf{q}}  \otimes  \hat{\bar{\mathbf{q}}}$, $\delta \bar{\mathbf{q}}\approx \begin{bmatrix} \frac{\delta{\bm \theta}}{2}^\intercal & 1 \end{bmatrix}^\intercal$.

A working assumption is that the prior map and global frame align along their $z$-direction~\cite{Dutoit2017ICRA} due to the observability of orientation in \ac{vins} up to rotations about the gravity vector. 
As such, the transformation only has four degrees of freedom: a relative position between the respective origins and a relative yaw. 
Parameterizing relative orientation using a single degree-of-freedom quaternion ---representing a rotation about a common gravity vector--- allows expressing the relative orientation state as
\[
{}_\mathsf{G}^\mathsf{M} \bar{\mathbf{q}} = 
\begin{bmatrix} 
 0 \\ 0 \\ \frac{\theta}{|\theta|}\sin{\frac{|\bm \theta|}{2}} \\ \cos{\frac{|\bm \theta|}{2}}
\end{bmatrix} \approx \begin{bmatrix} 
 0 \\ 0 \\ \frac{\delta \bm \theta}{2} \\ 1
\end{bmatrix} \otimes {}_\mathsf{G}^\mathsf{M} \hat{\bar{\mathbf{q}}}
\]
All updates to this quaternion are achieved by further rotations about the $z$-axis.

\subsection{Filter Propagation}

As the \ac{uav} moves through the environment, the \ac{imu}'s gyroscope and accelerometer collect measurements relating to the evolution of the \ac{imu} state. 
If $\bm \omega_m$ and $\mathbf{a}_m$ denote the local angular velocity and linear acceleration measurements while 
$\bm \omega$ and ${}^{\mathsf{I}}\mathbf{a}$ are the \emph{true} local angular velocity and local linear acceleration, ${}_\mathsf{G}^\mathsf{I}\mathbf{R}$ represents the rotation matrix associated with ${}^\mathsf{I}_\mathsf{G}\bar{\mathbf{q}}$, and with  $\mathbf{n}_w$ and $\mathbf{n}_a$ being continuous-time white Gaussian noise vectors that corrupt respective measurements, the
measurement model becomes
\begin{align*}
\bm \omega_m &= \bm \omega + \mathbf{b}_w+ \mathbf{n}_w \\
\mathbf{a}_m &= {}^\mathsf{I} \mathbf{a}- {}_\mathsf{G}^\mathsf{I}\mathbf{R} {}^G \mathbf{g}+ \mathbf{b}_a+ \mathbf{n}_a 
\end{align*}
Measurement biases are modeled as continuous-time random walks, driven by Gaussian white noises $\mathbf{n}_{bw}$ and $\mathbf{n}_{ba}$.
The dynamics of the state are now compactly expressed as
\[
\dot{\mathbf{b}}_w = \mathbf{n}_{bw}, 
\quad 
\dot{\mathbf{b}}_a = \mathbf{n}_{ba}, \quad 
\dot{\mathbf{x}}_\mathsf{static} = \mathbf{0},  
\quad {}_\mathsf{G}^\mathsf{I}\dot{\bar{\mathbf{q}}} = \tfrac{1}{2}\mathbf{\Omega}(\bm \omega) \;{}_\mathsf{G}^\mathsf{I}\bar{\mathbf{q}}
\]
\[
{}^\mathsf{G}\dot{\mathbf{p}} = {}^\mathsf{G} \mathbf{v}, \quad
{}^\mathsf{G}\dot{\mathbf{v}} = {}_\mathsf{G}^\mathsf{I} \mathbf{R}^\intercal \; {}^\mathsf{I} \mathbf{a}, \quad 
\mathbf{\Omega}(\bm \omega) = \begin{bmatrix} - \widehat{ \bm \omega } &  \bm \omega \\ -\bm \omega^\intercal & 0 \end{bmatrix} 
\]
while the target, letting $\mathbf{n}_T$ denote a Gaussian noise vector, is assumed evolving with $i \in \{0,\ldots,n-1\}$ as 
\begin{align*}
{}^\mathsf{G} \dot{\mathbf{p}}_T^{(i)} 
&= {}^\mathsf{G} \mathbf{p}_T^{(i+1)}\enspace, &
{}^\mathsf{G} \dot{\mathbf{p}}_T^{(n)} &= \mathbf{n}_T 
\end{align*}
giving rise to a linear target evolution model ---involving 
a $(3n+3)\times 3$ matrix $\mathbf{H}$ of all zeros and the last $3 \times 3$ block entry as the identity matrix, and a matrix $\mathbf{B}$ which is the target state's Jacobian--- and 
is of the form
\[
\dot{\mathbf{T}} = \mathbf{B}\, \mathbf{T} + \mathbf{H}\, \mathbf{n}_T
\]
%
%
%
%
%
Applying the expectation operator on these evolution equations, the dynamics of the state \textit{estimate}, $\hat{\mathbf{x}}$ (estimate notation $\hat{\cdot}$ should not be confused with the slightly wider differential geometric $\widehat{\cdot}$ wedge mapping) is
\[
{}_\mathsf{G}^\mathsf{I}\dot{\hat{\mathbf{q}}} 
= \tfrac{1}{2}\mathbf{\Omega}(\bm \omega_m -\hat{ \mathbf{b}}_{\omega}) {}_\mathsf{G}^\mathsf{I}\hat{\mathbf{q}} \enspace, 
\quad
 {}^\mathsf{G}\dot{\hat{\mathbf{v}}} 
 = {}_\mathsf{G}^\mathsf{I} \hat{\mathbf{R}}^\intercal 
 \left( \mathbf{a}_m-\hat{ \mathbf{b}}_{a} \right) + {}^\mathsf{G} \mathbf{g} 
\]
\begin{align*}
{}^\mathsf{G}\dot{\hat{\mathbf{p}}} 
&= {}^\mathsf{G} \hat{\mathbf{v}} \enspace, 
&
\dot{\hat{ \mathbf{b}}}_{\omega} 
&= \mathbf{0} \enspace, 
&
\dot{\hat{ \mathbf{b}}}_{a} &= \mathbf{0} \\
\dot{\hat{{\mathbf{x}}}}_\mathsf{static} &=  \mathbf{0} \enspace, &
\dot{\hat{\mathbf{T}}} &= \mathbf{B}\hat{\mathbf{T}} &
\end{align*}
These equations allow propagating the state estimate by analytically solving the differential equations across the measurement time-interval $\left[t_k, t_{k+1}\right]$.

In order to quantify the covariance of the propagated estimates, the state equations are linearized about the current estimates. 
With $\mathbf{F}$ and $\mathbf{G}$ being the Jacobians of the \ac{imu} state and noise dynamics respectively,  $\mathbf{n}_{imu}$ being the stacked vector of \ac{imu} noise, and $t$ and $m$ representing the dimensions of target and static states, respectively, this linearized dynamics takes the form
\begin{multline}
 \begin{bmatrix} 
 \delta\dot{{\mathbf{x}}}_\textsc{imu} \\   
 \delta\dot{{\mathbf{T}}} \\
 \delta\dot{{\mathbf{x}}}_\mathsf{static} 
  \end{bmatrix}  
  \approx 
  \begin{bmatrix} 
  \mathbf{F} & \mathbf{0}_{15 \times t} & \mathbf{0}_{15 \times m}  \\ 
  \mathbf{0}_{t \times 15} & \mathbf{B} &\mathbf{0}_{t \times m}  \\ 
  \mathbf{0}_{m \times 15} & \mathbf{0}_{m \times t} & \mathbf{0}_{t \times t} 
  \end{bmatrix} 
  \begin{bmatrix} 
  \delta{\mathbf{x}}_\textsc{imu}  \\ 
  \delta{\mathbf{T}} \\ 
  \delta{\mathbf{x}}_\mathsf{static} 
  \end{bmatrix}  \\ 
  + \begin{bmatrix} 
  \mathbf{G} & \mathbf{0}_{12 \times 3}  \\ 
  \mathbf{0}_{t \times 12} & \mathbf{H}  \\ 
  \mathbf{0}_{m \times 12} & \mathbf{0}_{m \times 3} 
  \end{bmatrix} 
  \begin{bmatrix} 
  \mathbf{n}_\textsc{imu} \\ 
  \mathbf{n}_T 
  \end{bmatrix}
\label{eq:err_ev}
\end{multline}

Let ${\bm \Phi}(t_{k+1},t_k)$ and ${\mathbf{A}}(t_{k+1},t_k) $ denote the state-transition matrices of the \ac{imu} and target from time $t_k$ to $t_{k+1}$ based on~\eqref{eq:err_ev}.
Matrices $\mathbf{Q}_\textsc{imu}$ and $\mathbf{Q}_{T}$ will denote the noise covariances for the \ac{imu}  and target state evolution, associated with discrete-time noise characterization~\cite{Maybeck1982, Trawny2005_Q_TR}.
After defining the following two matrices
\begin{align*}
{\bm \Psi} &= \begin{bmatrix} {\bm \Phi}(t_{k+1},t_k) & \mathbf{0}_{15 \times t} &\mathbf{0}_{15 \times m}   \\ \mathbf{0}_{t \times 15}  & {\mathbf{A}}(t_{k+1},t_k) &\mathbf{0}_{t \times m}   \\ \mathbf{0}_{m \times 15}  & \mathbf{0}_{m \times t}  & \mathbf{I}_{m \times m}  
  \end{bmatrix} \\
  \mathbf{Q}_k &= \begin{bmatrix} \mathbf{Q}_{imu} & \mathbf{0}_{15 \times t}  &\mathbf{0}_{15 \times m}   \\ \mathbf{0}_{t \times 15}  & \mathbf{Q}_{T} &\mathbf{0}_{t \times m}   \\ \mathbf{0}_{m \times 15}  & \mathbf{0}_{m \times t}  & \mathbf{0}_{m \times m} 
  \end{bmatrix}
\end{align*}
the propagation of the covariance $\mathbf{P}$ of the state estimator can be compactly expressed as
%
\[
 \mathbf{P}_{k+1} = {\bm \Psi}  \mathbf{P}_{k} {\bm \Psi}^\intercal + \mathbf{Q}_k
\]
\subsection{Filter Update}\label{sec:update}
%
Let ${}_\mathsf{I}^\mathsf{C}\mathbf{R}$ represent the (fixed) rotation matrix and ${}^\mathsf{C}\mathbf{p}_\mathsf{I}$ the displacement between the \ac{imu} and the camera.
Let ${}^\mathsf{G} \mathbf{p}_f$ be the pixel measurement associated with a 3D feature and set
\begin{equation}
 {}^{\sf c_k} \mathbf{p}_f = {}_\mathsf{I}^\mathsf{C}\mathbf{R}\; 
 {}_\mathsf{G}^{\mathsf{I}_k}\mathbf{R} \;
 ({}^\mathsf{G} \mathbf{p}_f - {}^\mathsf{G} \mathbf{p}_{\mathsf{I}_k})+ 
 {}^\mathsf{C}\mathbf{p}_\mathsf{I} \label{bearing}
\end{equation}
Expand
$
{}^\mathsf{c_k}\mathbf{p}_f = \begin{bmatrix}
{}^{\sf c_k}x_f &
{}^{\sf c_k}y_f &
{}^{\sf c_k}z_f
\end{bmatrix}^\intercal
$
and define the map
\begin{equation}
\Pi \left( {}^\mathsf{c_k}\mathbf{p}_f \right) =\begin{bmatrix} ({}^{c_k}x_f)/({}^{c_k}z_f) \\ 
({}^{c_k}y_f)/({}^{c_k}z_f)
\label{eq:bearing}
\end{bmatrix}
\end{equation}
The pixel measurement is assumed to have noise $\mathbf{n}_f$.
This measurement 
projected into a camera associated with timestep $k$ is generated by a measurement function 
\[
\mathbf{z}_k =
\begin{bmatrix} u_k \\ v_k \end{bmatrix} = \mathbf{h}(\mathbf{x}) + \mathbf{n}_f = 
\Pi ({}^{\sf c_k} \mathbf{p}_f) + \mathbf{n}_f 
\]
 This function transforms the 3D position of the feature into the camera's local frame, and then  projects it onto the image plane. Similarly, the projection of the target into the plane can be described by replacing the unknown feature position in \eqref{eq:bearing} with the target's position. 
A similar procedure is followed for a known landmark\footnote{This is assumed perfectly known, no noise. To include this noise into the system, one could use the computationally efficient Cholesky-Schmidt-Kalman Filter proposed in~\cite{Dutoit2017ICRA}. }  expressed in the (inertial) map frame, ${}^\mathsf{M} \mathbf{L}_f$, to the current camera frame,
\[
  {}^{\sf C_k}\mathbf{L}_f = {}^\mathsf{C}_\mathsf{I}\mathbf{R}\; 
  {}^{\mathsf{I}_k}_\mathsf{G}\mathbf{R}\; 
  \left( {}^\mathsf{G} \mathbf{p}_\mathsf{M} + {}^\mathsf{M}_\mathsf{G} \mathbf{R}^\intercal \; {}^\mathsf{M}\mathbf{L}_f- {}^\mathsf{G}\mathbf{p}_{\mathsf{I}_k} \right) + {}^\mathsf{C} \mathbf{p}_{\sf I} 
\]
resulting in a projected measurement 
\[
    \mathbf{z}_f = \Pi({}^{\mathsf{C}_k}\mathbf{L}_f) + \mathbf{n}_f 
\]
These three different bearing measurement types define the filter updates. For the target and map localization measurements, a standard \ac{ekf} update can be utilized ---all the quantities involved in the measurement are contained in the state.  

At this stage, in order to use the unknown feature bearing measurements, linear marginalization through the \ac{msckf} update step~\cite{Mourikis2007ICRA} is performed, since otherwise 
a standard \ac{ekf} storing all the measured features in the state vector would lead to unbounded computation. 
Let $\mathbf{H}_x$ and $\mathbf{H}_f$ be the measurement Jacobians with respect to the estimated state and the unknown feature, and $\delta{\mathbf{x}}$ and $\delta{\mathbf{p}}_f$ are the corresponding state errors, while $\mathbf{n}_f$ denotes the stacked measurement noise.
Given a vector $\mathbf{z}$ of pixel measurements associated with a feature that has been tracked across the window, and the corresponding stacked measurement generation function $\mathbf{h}(\mathbf{x})$, the linearized measurement residual $\tilde{\mathbf{z}}= \mathbf{z}- \mathbf{h}(\hat{\mathbf{x}})$ is expressed as
\[
\tilde{\mathbf{z}} = \mathbf{H}_x \; \delta{\mathbf{x}} + \mathbf{H}_f \; \delta{\mathbf{p}}_f + \mathbf{n}_f
\]
 The linearization point for the feature position is found through triangulation using the corresponding bearing measurements and the current state estimates. 
By performing QR decomposition on $\mathbf{H}_f$, it is possible to define the matrix $\mathbf{Q}_2$ whose columns span the left nullspace of $\mathbf{H}_f$. 
If $\mathbf{R}$ denotes the covariance of the original stacked measurement 
and 
\begin{align*}
\mathbf{n}' &\sim \mathcal{N}\left(\mathbf{0}, \mathbf{R}' \right) \enspace, &
\mathbf{R}' &=  {\mathbf{Q}_2}^\intercal \;\mathbf{R} \; \mathbf{Q}_2
\end{align*}
then multiplying by ${\mathbf{Q}_2}^\intercal$ \textit{removes} the dependency on feature errors
\[
{\mathbf{Q}_2}^\intercal \; \tilde{\mathbf{z}} = {\mathbf{Q}_2}^\intercal \; \mathbf{H}_x \, \delta{\mathbf{x}} +  {\mathbf{Q}_2}^\intercal \;\mathbf{n}_f 
\Rightarrow \tilde{\mathbf{z}}' = \mathbf{J}_x \; \delta{\mathbf{x}} + \mathbf{n}'
\]%
The \ac{msckf} therefore creates a new residual, $\tilde{\mathbf{z}}'$, with corresponding measurement Jacobian, $\mathbf{J}_x$, that does not require storing features into the state vector, as the transformed measurement relates only to the current \ac{imu} state and the clones in the window. 

\subsection{State Initialization}

\subsubsection{Map Initialization}

As the filter does not start with a prior estimate of the map transformation parameters, it is important to perform initialization of these quantities. In particular,
bearing measurements to known landmarks are collected until the transformation between the global and prior map frames can be estimated. An estimate of the landmarks in the \ac{imu}'s global frame, ${}^\mathsf{G} \mathbf{L}_f$, can be recovered through multi-view triangulation. Using the positions for the landmarks expressed in both frames, the 6 DOF relative pose is estimated~\cite{Horn1987JOSAA}, from which the relative position and yaw are extracted. 
If $\tilde{\mathbf{z}}_\textsf{MI}$ denotes the stacked residual associated with these bearing measurements and $\mathbf{n}_\mathsf{MI}$ the stacked measurement noise, then  using this estimate the linearized system for the landmarks' bearing measurements is built
\begin{equation}
\tilde{\mathbf{z}}_\textsf{MI} = \mathbf{H}_x \;\delta \mathbf{x} + \mathbf{H}_M \; \delta \mathbf{x}_\textsc{map} + \mathbf{n}_\mathsf{MI} \label{map_init}
\end{equation}
%
The Jacobians $\mathbf{H}_x$ and $\mathbf{H}_M$ have been separated with respect to the already initialized state and the new map parameters, respectively. Using delayed initialization these new map parameters can be added to the state. All future bearing measurements to known landmarks can now be processed using a standard \ac{ekf} update. 

\subsubsection{Target Initialization}
The target is initialized in a similar manner by collecting $N$ bearing measurements corresponding to times $\{t_0, t_1, \cdots, t_N\}$, where $t_0$ denotes the first time that the target is seen.  
Let ${}^\mathsf{G}\mathbf{p}_{T_0}^{(i)}$, ${}^\mathsf{G}\mathbf{p}_{c_j}$,  $d_j$, $\mathbf{b}_j$, $\Delta t_j= t_j-  t_0$ be the initial $i$-th derivatives of the target's position, the position of the \ac{uav}'s measuring camera (left or right) in the global frame at time $t_j$, the depth of the target in the j-th image, the measured bearing vector from the \ac{uav} to the target expressed in the global frame, and the time since the initial target measurement, respectively.
Assuming a constant derivative model over the interval, the constraint satisfied at time $t_j$ is given by:
%
\[
{}^\mathsf{G}\mathbf{p}_{\sf c_j} = -d_j \, \mathbf{b}_j +\sum_{i=0}^n \frac{(\Delta t_j)^i}{i!}\;{}^\mathsf{G}\mathbf{p}_{T_0}^{(i)}
\]
Construct $\mathbf{p}$ by stacking all the relevant camera poses, and let $\mathbf{T}_0$ be the initial target state.
Set $\mathbf{d}$ to be the vector of target depths, and let $\mathbf{Y}$ be the matrix encoding the constraints.
Stacking the equations and separating out the unknowns leads to the following least-squares solution
%
%
\[
\mathbf{p} = \mathbf{Y} \begin{bmatrix} \mathbf{T}_0 \\ \mathbf{d} \end{bmatrix} \Rightarrow \begin{bmatrix} \hat{\mathbf{T}}_0 \\ \hat{\mathbf{d}} \end{bmatrix} = 
\left(\mathbf{Y}^\intercal \; \mathbf{Y} \right)^{-1} \mathbf{Y}^\intercal \, \mathbf{p} 
\]
%
%
%
Propagating this initial state yields estimates of the target at each of the measuring times, $\{\hat{\mathbf{T}}_0, \hat{\mathbf{T}}_1, \cdots,  \hat{\mathbf{T}}_N\}$. In order to add the target states into the filter, the motion model is treated as a measurement constraining the target states between two consecutive times. For example, between timesteps $k$ and $k+1$, for $\mathbf{n}_{TD} \sim \mathbf{N}
\left(\mathbf{0}, \mathbf{Q}_T \right)$,
\[
\mathbf{0} = -\mathbf{T}_{k+1}+ \mathbf{A}\left(t_{k+1},t_k \right)\mathbf{T}_{k} + \mathbf{n}_{TD} 
\]
The target evolution and target bearing measurements are stacked into a single vector and used to perform delayed initialization as in~\eqref{map_init}.
The filter then will contain estimates for the target at each of the imaging times, and so all but the active target state are marginalized. 

\section{Simulations}

\subsection{Estimation Validation}

\begin{figure}
	\centering
	\subfigure[]{
		\includegraphics[width=0.3\textwidth]{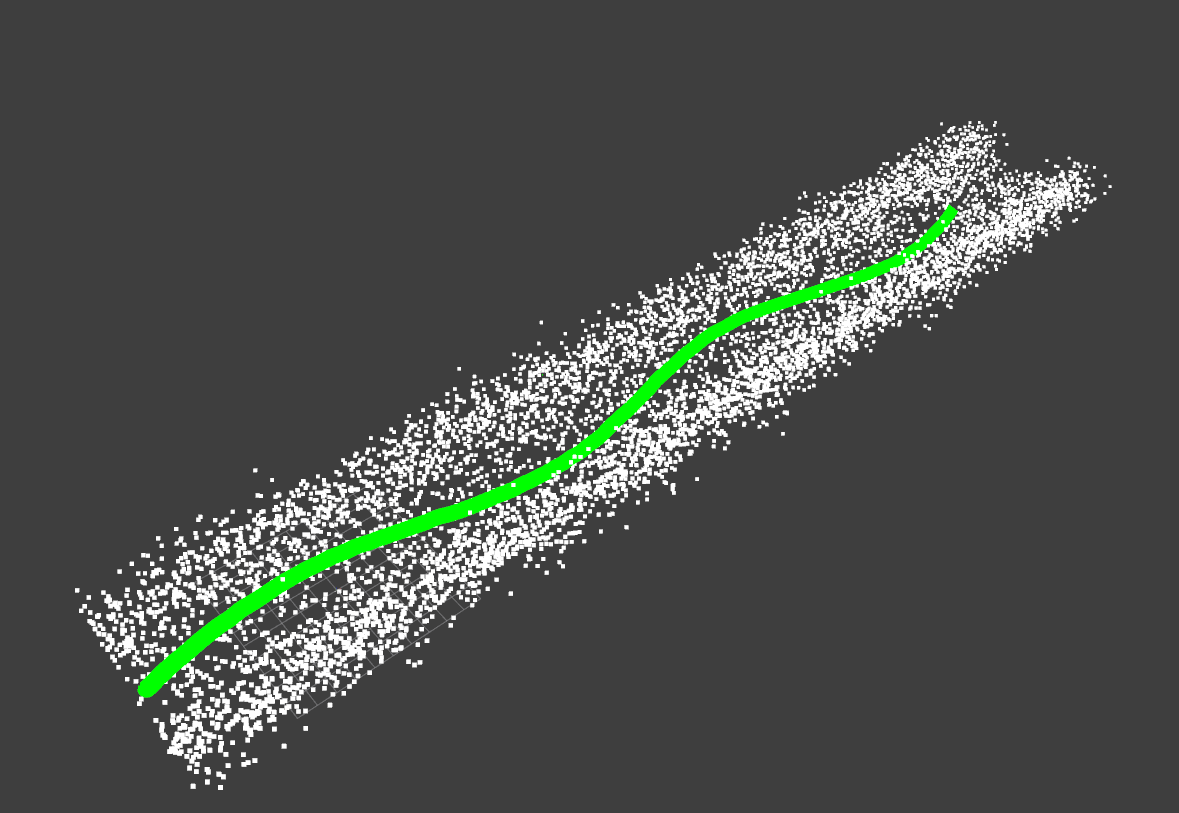} 
        \label{fig:hallway}
    }
	\subfigure[]{
	    \includegraphics[width=0.5\textwidth]{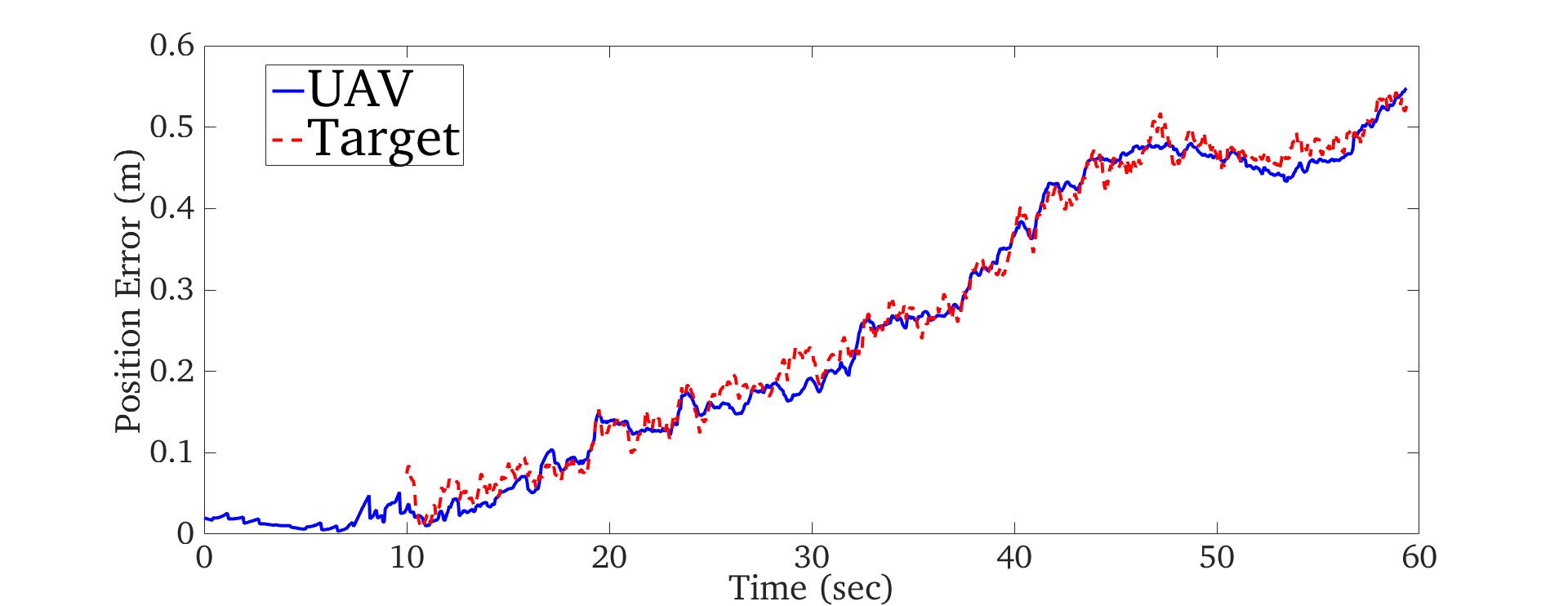}
        \label{fig:normed_position_error}
	}
	\caption{Estimation validation: \subref{fig:hallway} Hallway simulated to validate estimation and target tracking with the \ac{uav} trajectory shown in green. \subref{fig:normed_position_error} The normed position error in the hallway scenario. The \ac{uav} is able to accurately estimate its state and that of the target even without relying on landmark estimates. 
		\label{Fig:est}
    }
	
\end{figure}

Being a significant and new addition, the estimation module of the system is first validated using simulated sensors from the ROS/Gazebo package \texttt{RotorS}~\cite{Furrer2016}, without incorporating motion-planning or in-house developed controllers. 
In a simulated environment, a Firefly \ac{uav} equipped with a forward facing \textsc{vi}-sensor executes a sinusoidal trajectory of approximately 60\,m at a speed of 1\,m/s.  
A Pelican \ac{uav}, flying in front of the Firefly at constant velocity, serves as the target, and the estimator assumes this target to follow a constant velocity motion model. 
Unknown features are simulated along the walls and floor of the hallway (Fig.~\ref{fig:hallway}). 
This test is performed {\em without} known landmarks, in order to show that the proposed estimator also works in unmapped environments. 
Over a path of length 58\,m, the robot position estimate has a \ac{rmse} and ending error of 0.2379\,m and 0.55\,m, respectively, 
while the target position \ac{rmse} and ending error are 0.2835\,m and 0.52\,m (Fig.~\ref{fig:normed_position_error}).


\subsection{System Validation}

\begin{figure} 
	\centering
	\subfigure[]{
		\includegraphics[width=.8\columnwidth]{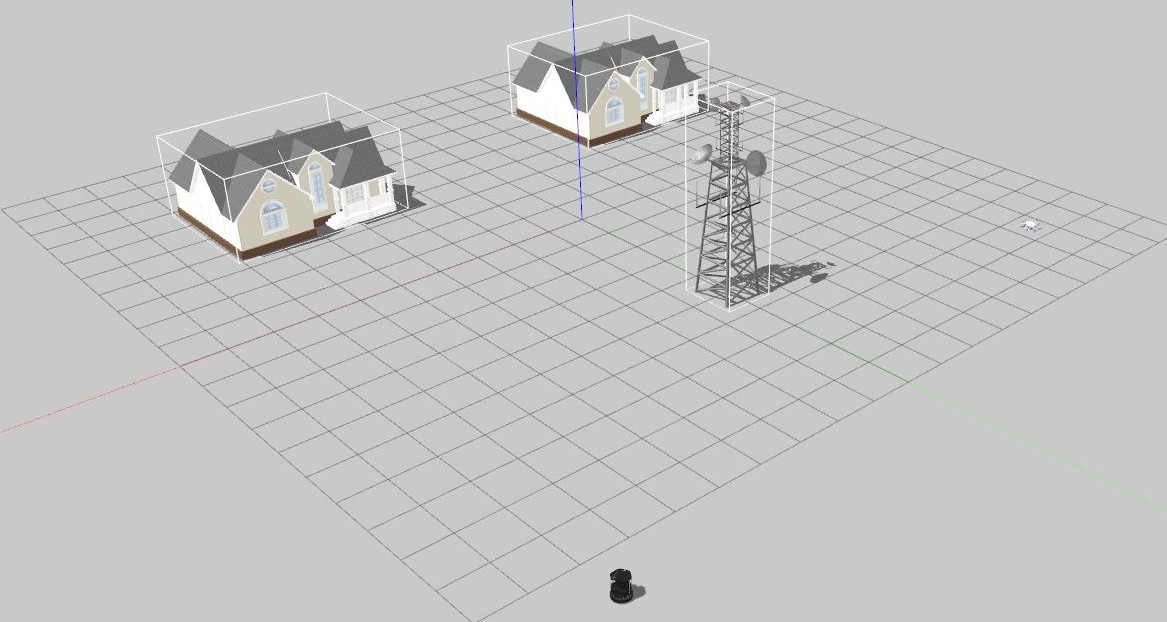} 
        \label{workspace}
    }
	\subfigure[]{
	    \includegraphics[width=.7\columnwidth]{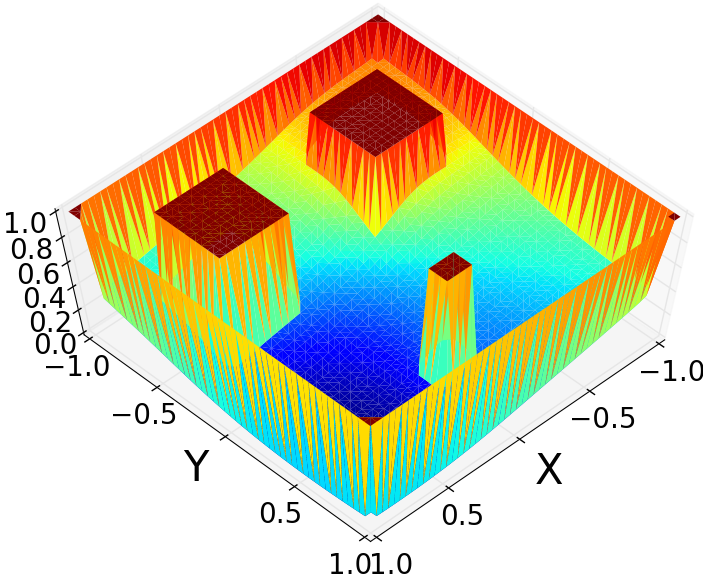}
        \label{nav_func}
	}
	\caption{System validation: \subref{workspace} The simulated workspace whose boundary is not shown for clarity. \subref{nav_func} Navigation function constructed on unit squircle world in the plane containing the target at location $(7.5\text{m},\enspace 5\text{m}, \enspace 0.5\text{m})$. Variation across different heights can be seen in the attached video. 
		\label{Fig:system}
    }
	
\end{figure}

The workspace in which the simulated detection scenario evolves is shown in Fig.~\ref{workspace}. 
There are three (interior) obstacles in this workspace: two houses and a radio tower (not to scale). 
The workspace boundary (not shown in Fig.~\ref{workspace}) is modeled as a squircle (rounded box) of dimension $20\,\text{m}\times20\,\text{m}\times5\,\text{m}$ centered at coordinate $(0, \, 0, \, 2.5)$ ---all components in m; the interior obstacles are also modeled as squircles of dimension equal to their bounding boxes. 
Figure~\ref{nav_func} shows a section of the constructed navigation function with  a target location at coordinate $(7.5, \, 7.5, \, 0.5)$. 
(The variation of the function in the $z$ direction for the same target location can be seen in the video submitted.)
The navigation function parameters for this setup were $\kappa = 6$, $\lambda = 10^{4}$ and $\lambda_\mathrm{sq} = 10^{4}$. 
The gradient of the function gives a desired velocity orientation for the \ac{uav}, while the yaw angle is chosen so that the \ac{uav} always faces the estimated target location. 

A dense set of 1550 unknown features and a sparse set of 160 known landmarks is randomly placed over different surfaces in the environment. 
The estimator subscribes to the ground-truth \ac{uav} and target odometry, as well as the noisy \ac{imu} at 200\,Hz, while publishing the resulting \ac{uav} and target estimates at the \ac{imu} rate. 
Bearing measurements are simulated at a rate of 20\,Hz, and are corrupted by simulated noise following a Gaussian distribution with one pixel standard deviation. 
These measurements are then processed using the proposed algorithm with an \ac{msckf} window size of 8. 
Occlusions are also simulated by checking whether each bearing ray intersects any of the obstacle boundaries, which are approximated as rectangular boxes.


In this scenario, a \ac{uav} (Firefly) equipped with a $20^\circ$ downward facing \textsc{vi}-sensor intercepts a target (which is realized by a turtlebot) assumed to carry an isotropic Cf-252 neutron source. 
Radiation emitted by this source is assumed to be picked up by a Domino thermal neutron detector\footnote{Radiation Detection Technologies Inc {\color{blue}http://radectech.com/products/rdt-domino-v5-4}} assumed mounted on the \ac{uav}. 
The radioactivity is simulated using the thining algorithm~\cite{Pasupathy} ---benchmarked against a $5\,\mu$Ci (micro Curie) source available--- to guide the simulated realization of higher activity sources.
(The activity of a $5\,\mu$Ci source, 1.5\,m away from the target essentially blends completely into background.)

The case of {\sf Maneuver-1} in the companion video\footnote{ \color{blue}\url{https://www.youtube.com/watch?v=zur04_avhJg}} illustrates the  motion planning approach while {\sf Maneuver-2} showcases a sharp turn by the \ac{uav} to avoid an obstacle. 
The results reported below refer exclusively to {\sf Maneuver-1}. 
There, the target moves in a straight line with a velocity of 1\,m/s and the \ac{uav} intercepts it with a maximum speed of 6\,m/s. 
The motion planning process consists of three phases: 
\begin{inparaenum}[(i)]
\item hovering at 2\,m using a position controller to initialize the tracking, 
\item target chase using the navigation function based planner and velocity controller, and 
\item hovering again using the position controller once the target crosses the detection area. 
\end{inparaenum}

\begin{figure*}[thbp]
	\vspace{-0.1cm}
	\centering
	\subfigure[]{
		\includegraphics[height=.2\textwidth,width=0.3\linewidth]{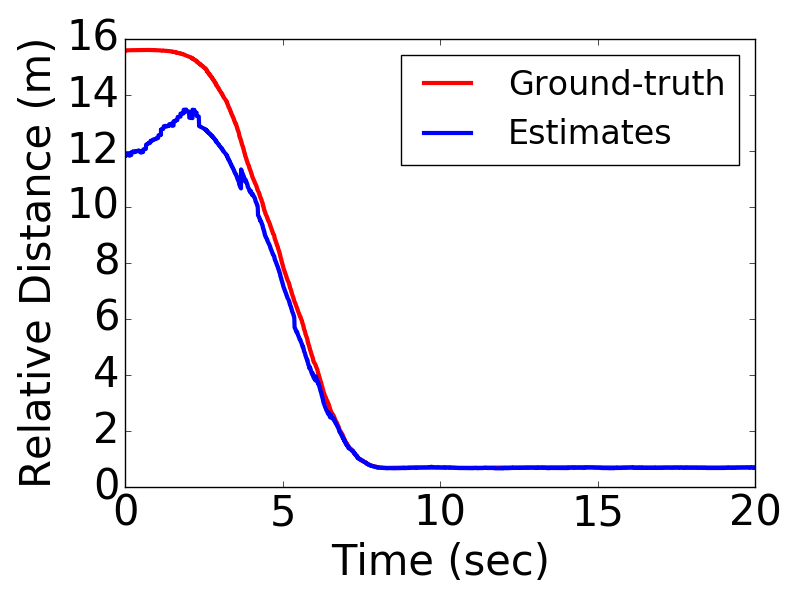} 
        \label{relative_distance}
    }
	\subfigure[]{
	    \includegraphics[height=.2\textwidth,width=0.3\linewidth]{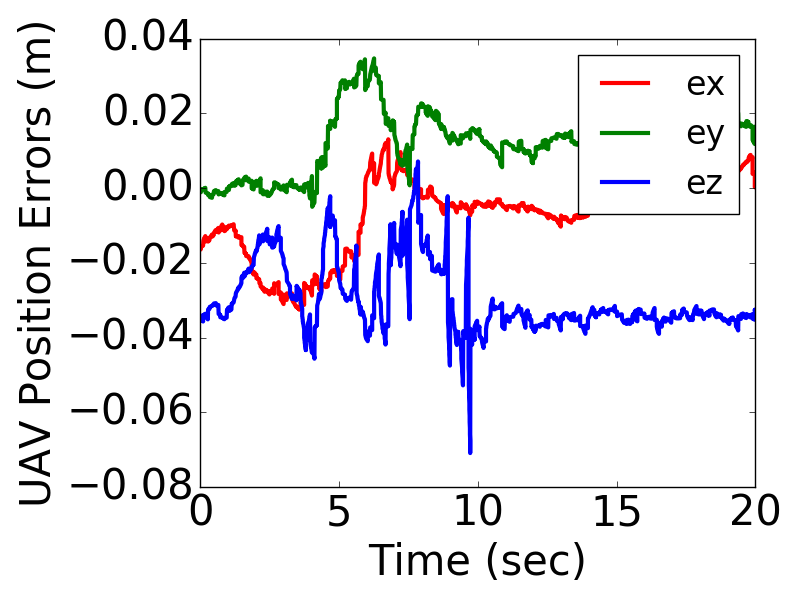}
        \label{robot_estimation_error}
	}
    \subfigure[]{
    \includegraphics[height=.2\textwidth,width= .3\textwidth]{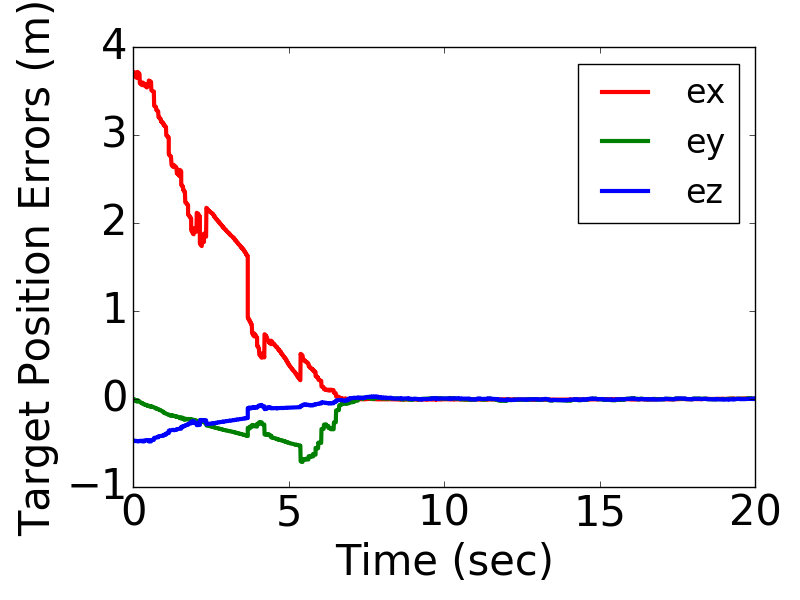}
    \label{target_estimation_error}
    }
	\caption{\subref{relative_distance} Estimated and Ground-truth values of Relative distance between the \ac{uav} and the target. \subref{robot_estimation_error} Error in VINS position estimates of the \ac{uav}. \subref{robot_estimation_error} Corresponding errors for the target.
    }
\end{figure*}

The \ac{uav} starts chasing the target at 6\,m/s, and eventually follows the target maintaining a constant distance of about 0.7\,m. 
The true and estimated relative distance between the \ac{uav} and the target has been shown in Fig.~\ref{relative_distance}. 
The error in these two eventually converges to approximately 15--20\,mm.  
%
Figure \ref{robot_estimation_error} and \ref{target_estimation_error} shows the $x$, $y$ and $z$ components of the error between estimates, and ground truth positions of the \ac{uav} and the target. 
The estimation error in the target's position starts with a high value due to initialization at a far distance, but eventually converges to acceptable values. The \ac{uav} state estimation error, due to access to highly informative known landmark bearing measurements, remains bounded within 10\,cm over the path. 


\begin{figure*}[thbp]
	\centering
    \subfigure[]{
		\includegraphics[height=.2\textwidth,width=0.3\linewidth]{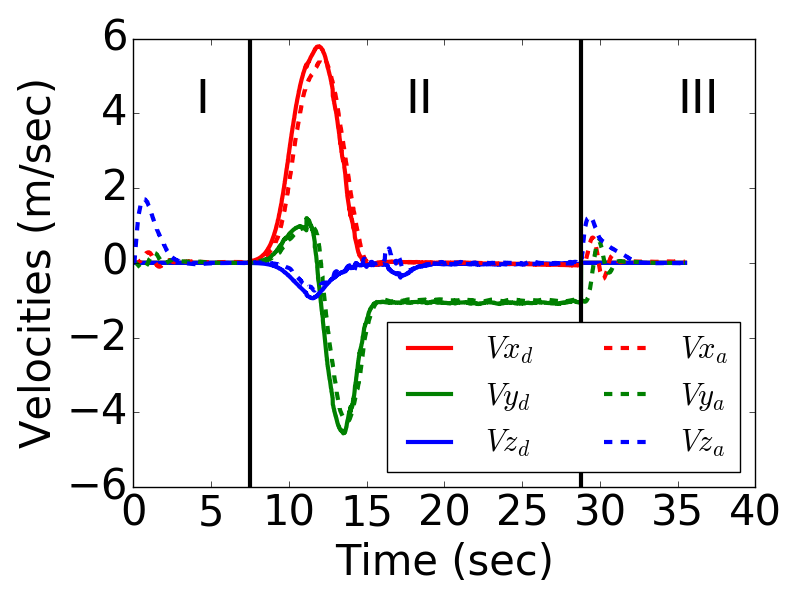} 
        \label{velocities}
    }
	\subfigure[]{
		\includegraphics[height=.2\textwidth,width=0.3\linewidth]{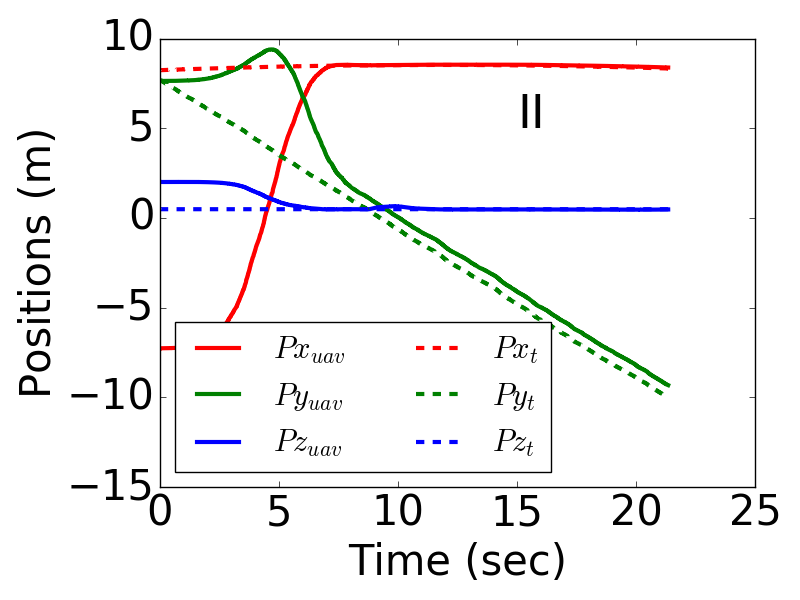} 
        \label{positions}
    }
	\subfigure[]{
	    \includegraphics[height=.2\textwidth,width=0.3\linewidth]{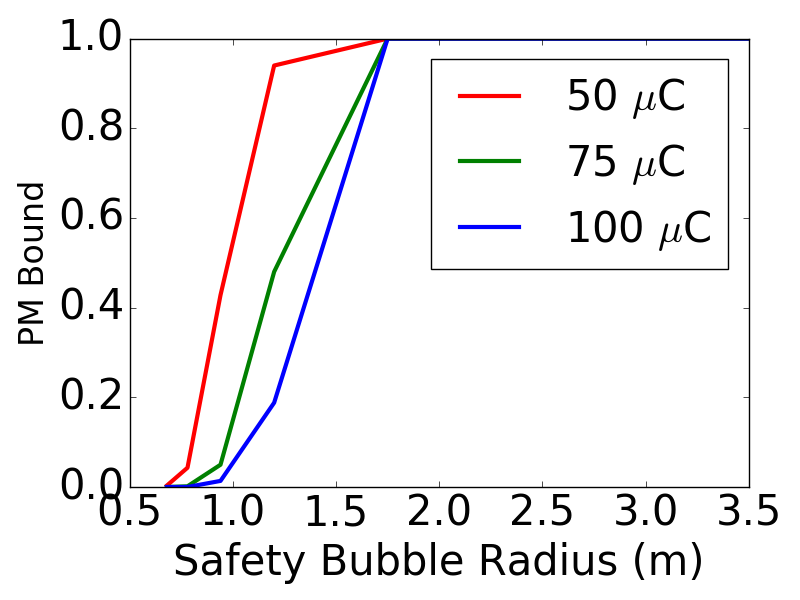}
        \label{bound_on_PM}
	}
	\caption{\subref{velocities} Desired and achieved velocities in various phases of \ac{uav} motion. \subref{positions} Achieved positions of the \ac{uav} and the target in the chase phase (II). \subref{bound_on_PM} Bound on Probability of missed detection with varying minimum distance between the robot and the target; calculated for the $T=$ 20 seconds.
		\label{PM}
    }

\end{figure*}

Figure~\ref{velocities} presents the desired and achieved velocities of the \ac{uav}. 
The fluctuations in achieved velocities at the start of phases (i) and (iii) are due to commanding the \ac{uav} to hover. Figure~\ref{positions} shows the $x$, $y$ and $z$ components of the trajectory of the \ac{uav} and the target during phase (ii), which have been used in subsequent radiation detection calculations. 
%
%
%

Figure \ref{bound_on_PM} justifies tightly coupling state estimates to the target tracking as well as using a navigation function based planner. 
It presents the variation of the bound on probability of missed detection with the radius of safety bubble around the target, for a bound on probability of false alarm $\alpha = 0.001$. Multiple trajectories during the chase phase were generated with different safety bubble radii around the target to generate Fig \ref{bound_on_PM}. Color-coded lines show bounds calculated assuming source strengths of $50 \mu$Ci, $75 \mu$Ci, and $100 \mu$Ci, for which the activity $a$ is $2.2\times 10^{5}$, $3.3\times 10^{5}$, and $4.4\times 10^{5} \ac{cps}$ respectively.\footnote{The value of source activity is calculated assuming $4\times 10^{9}$ neutrons per Curie per second emitted by Cf-252 source. (See {\color{blue}{http://www.logwell.com/tech/nuclear/Californium-252.html}}).} 

The background radioactivity and detector's radiation cross-section used in the calculations were $0.005833 \ac{cps}$  and $2.12\times 10^{-6} \text{m}^2$. 
The total time allocated to make the decision was capped at $T = 20$ sec. 
The value of bound on probability of missed detection equal to 1 indicates the inability to make any accurate decision. 
It can be seen that \ac{uav} needs to be within 1--1.5\,m radius of the target in order to detect a low intensity source using \ac{cots} detectors, guiding the choice of the proposed approach. 

\section{Conclusions and Future Work}

An integrated motion-planning framework that tightly-couples visual-inertial navigation and target tracking with a 3D navigation function-based planner and geometric controller is shown to be appropriate for applications in \ac{uav}-based radiation detection.
Such an integrated system offers close tracking for accurate radiation detection, while avoiding collisions with obstacles in the environment. 
Future effort will be directed towards extending this methodology to more realistic and efficient scenarios such as monocular camera deployments with noisy prior landmarks, handling more complicated target motion models, experimentally validating the system, and investigating the effect of navigation function tuning on the \ac{uav} maneuvers.

\bibliographystyle{IEEEtran}
\bibliography{Effect_of_noise_on_radiation_detection_R0,DecisionMaking,decision_graph,Networking,vins_library,extra} 

\end{document}